%% file: main.tex
\documentclass[runningheads]{llncs}

\usepackage{eccv}

\usepackage{eccvabbrv}

\usepackage{graphicx}
\usepackage{booktabs}

\usepackage[accsupp]{axessibility}  

\usepackage{hyperref}

\usepackage{orcidlink}

\usepackage{multirow}
\usepackage{makecell}
\usepackage[most]{tcolorbox}

\usepackage{tabularx}
\usepackage{adjustbox}
\usepackage{makecell}
\usepackage{booktabs}
\usepackage{enumitem}

\newcommand{\Ours}{Goku\xspace} 
\newcommand{\OurData}{\Ours}   
\newcommand{\OurBench}{\Ours-Bench\xspace} 
\newcommand{\OurModel}{\Ours-Edit\xspace}       

\usepackage{booktabs}
\usepackage{pifont}
\usepackage[table]{xcolor} 
\newcommand{\cmark}{\textcolor{green!60!black}{\ding{51}}}
\newcommand{\xmark}{\textcolor{gray}{\ding{55}}}
\usepackage{wrapfig}  

\newcommand{\blfootnote}[1]{%
  \begingroup
  \renewcommand\thefootnote{}\footnotetext{#1}
  \endgroup
}

\begin{document}

\title{\OurData: A Million-Scale Universal Dataset and Benchmark for Instruction-Based Video Editing} 

\titlerunning{\OurData}

\author{Sen Liang\inst{1,2}$^{\star}$ \and
Cong Wang\inst{2}$^{\star}$ \and
Zhentao Yu\inst{2} \and
Fengbin Guan\inst{1} \and
Zhengguang Zhou\inst{2} \and
Teng Hu\inst{2} \and
Youliang Zhang\inst{2} \and
Yuan Zhou\inst{2} \and
Xin Li\inst{1} \and
Qinglin Lu\inst{2} \and
Zhibo Chen\inst{1}$^{\dagger}$}

\authorrunning{S. Liang et al.}

\institute{University of Science and Technology of China, Hefei, China\\
\and
Tencent Hunyuan, China\\
\email{liangsen@mail.ustc.edu.cn, chenzhibo@ustc.edu.cn} 
\url{http://flying-sky999.github.io/Goku.github.io/}}



\maketitle
\blfootnote{$^{\star}$ Equal contribution.  \quad $^{\dagger}$ Corresponding author.}

\input{0_abstract}

\input{1_intro}

\input{2_related_work}

\input{3_method}

\input{3_5_bench}
\input{4_experiment}
\input{5_summary}




%
%
\bibliographystyle{splncs04}
\bibliography{main}

\end{document}

%% file: 0_abstract.tex
\begin{figure*}[t]
  \centering
    \includegraphics[width=1\linewidth]{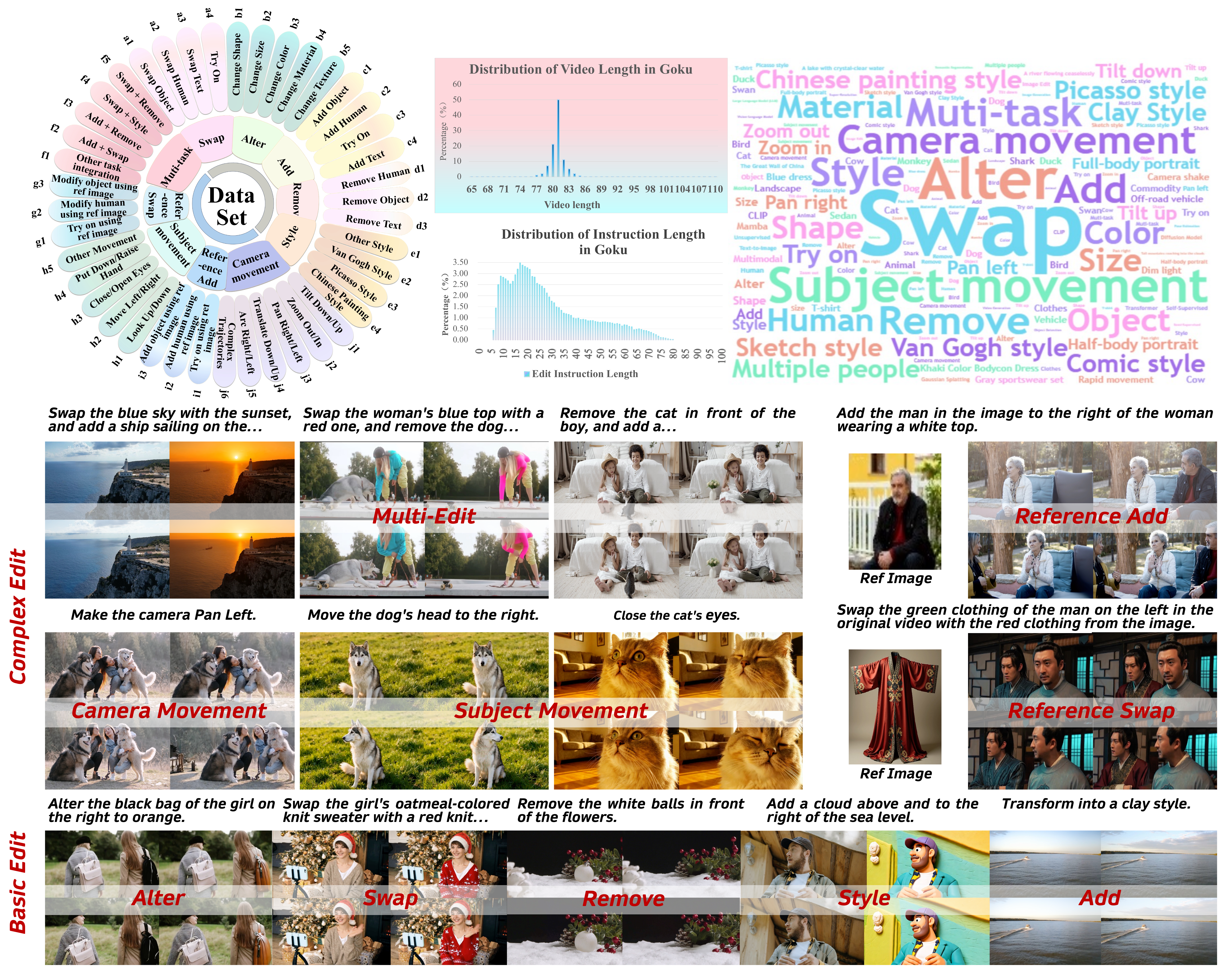}
    \caption{\OurData covers 10 core video editing task classes across basic and complex edits. The word cloud illustrates the instruction vocabulary distribution, while the two charts show the distributions of instruction length and frame count.}
  \label{fig:2}
\end{figure*}

\begin{abstract}

Existing instruction-based video editing datasets commonly focus on single-task appearance editing, failing to meet the complex creative demands of real-world scenarios.
To bridge this gap, we present \OurData, a large-scale dataset featuring 2 million high-quality, instruction-aligned video editing pairs, which is the first to extend task boundaries from basic appearance editing to multi-task and structural manipulations (e.g., precise control of subject movement).
To tackle the data synthesis challenges inherent in these complex tasks,
we design an efficient data synthesis pipeline that decomposes complex edits into controllable sub-problems and introduce a progressive filtering system for data reliability throughout the whole process.
Furthermore, we explore the optimal network structures on \OurData, and propose \OurModel. To deeply comprehend complex editing instructions, \OurModel leverages an MLLM as its text encoder and adopts a decoupled dual-branch design: a dedicated mask branch handles structural control, freeing the main branch for appearance rendering. 
A comprehensive video editing benchmark, \OurBench, is also proposed with 1,000 human-verified test cases and 7 novel editing-specific metrics.
Evaluated on \OurBench, \OurModel obtains up to +8\% improvement on other open-source models in terms of instruction following.

\end{abstract}

%% file: 1_intro.tex
\section{Introduction}
\label{sec:intro}

The rapid development of generative AI has fundamentally reshaped the landscape of digital content creation, transitioning from simple video synthesis~\cite{bar2024lumiere,zhou2024allegro,svd,ayl,guo2023animatediff,zhou2022magicvideo,walt,wang2023modelscope,vdm,videogan,cvideogan,singer2022make,text2video,lin2025apt,ma2024follow, liang2026spongebob, liang2025omniv2v}  to the more ambitious domains.
In this context, instruction-based video editing (IVE)~\cite{InsV2V,Insvie-1m, ku2024anyv2v, StableV2V, tan2025omnivideo, liang2026cot} proposes synthesizing video triplets, a paradigm that offers an intuitive and flexible interface for users, democratizing professional-grade video production. 
However, despite the impressive capabilities demonstrated by state-of-the-art models, they remain largely confined to single-task and appearance-level modifications, such as object removal and single-attribute alteration.

One of the primary factors contributing to this limitation is the narrow scope of existing problem definitions within current datasets~\cite{InsV2V, Insvie-1m, zi2026senorita, ditto, OpenVE}. 
These datasets often oversimplify editing tasks, neglecting the critical need for \textbf{complex structural transformations} and \textbf{simultaneous multi-task editing} inherent in real-world scenarios.
For instance, Ditto~\cite{ditto} improves quality control but is heavily skewed toward style transfer, limiting task diversity. Similarly, the concurrent OpenVE-3M~\cite{OpenVE} tightens filtering rigor yet remains confined to appearance-level edits, leaving structural and multi-task editing unaddressed.
Consequently, there is still a pressing need for a more comprehensive dataset that can bridge the gap between simplistic attribute changes and the multifaceted requirements of practical video manipulation.

Therefore, we introduce \OurData, a large-scale, comprehensive dataset comprising 2 million high-quality, instruction-aligned video editing pairs, as shown in Fig.~\ref{fig:2}.
Unlike previous datasets that treat editing as a collection of isolated tasks, \OurData is designed to encompass a diverse spectrum of challenges.
As shown in Tab.~\ref{tab:data_compare}, our framework not only covers fundamental single-task and appearance-level edits (basic edits) but also, for the first time, explicitly incorporates multi-task (i.e., Multi-Task Edit), structural deformations (i.e., Camera Movement and Subject Movement), and reference-guided editing tasks.

To construct such a vast and multifaceted dataset, we develop a scalable and automated data generation pipeline designed to ensure both semantic precision and temporal coherence.
Specifically, for basic appearance-level editing (e.g., Swap), we follow established practices by employing VACE~\cite{jiang2025vace} for robust data synthesis;
for structural editing and multi-task editing, we decompose complex instructions into independently controllable sub-problems and tackling them with task-specific expert models. To prevent error accumulation during the sub-problem cascade and ensure consistency across editing steps, we introduce a rigorous progressive filtering system powered by Gemini2.5-Pro. At each critical stage of the pipeline, entries are filtered across three dimensions: instruction alignment, frame-to-frame stability, and perceptual photorealism. This ensures that \OurData provides the highest-fidelity data for the community.

Leveraging the rich annotations and complex scenarios within our dataset, we develop Goku-Edit, a robust framework specifically optimized for multifaceted video manipulation.
To effectively parse and execute complex editing instructions, we follow the recent paradigm of employing a Multimodal Large Language Model (MLLM) as the text encoder and incorporate an additional mask prediction branch that serves to enhance the model's spatial grounding, allowing the main branch to focus more on fine-grained appearance details.
By coupling MLLM-based reasoning with fine-grained spatial constraints, Goku-Edit can accurately perform sophisticated edits that were previously unattainable.

Finally, to ensure a rigorous assessment, we establish \OurBench, a comprehensive benchmark featuring 1,000 diverse test cases and 7 specialized metrics tailored for complex video editing tasks.
All test cases and evaluation metrics have undergone meticulous human verification to ensure their reliability and fairness. 
Extensive experiments conducted on Goku-Bench demonstrate that our Goku-Edit achieves up to an 8\% improvement over existing state-of-the-art models in instruction following.

In summary, our main \textbf{contributions} are summarized as follows:
\begin{itemize}
\item[\textbf{(1)}] \textbf{A Comprehensive Dataset:} we introduce \textbf{\OurData}, the most extensive IVE dataset to date, featuring 2 million high-quality video pairs that cover complex structural and multi-task editing for the first time.
\item[\textbf{(2)}] \textbf{A Robust Data Pipeline:} we design a scalable data pipeline with a progressive filtering system for semantic precision and temporal coherence.
\item[\textbf{(3)}] \textbf{A Versatile Editing Model:} we propose \textbf{\OurModel}, which effectively bridges high-level semantic reasoning with precise spatial manipulation via an MLLM-based text encoder and a novel mask prediction branch.
\item[\textbf{(4)}] \textbf{A Rigorous Benchmark:} we establish \textbf{\OurBench}, a human-verified benchmark with 1,000 diverse cases and seven specialized metrics, providing a new standard for evaluating sophisticated video editing models.

\end{itemize}

%% file: 2_related_work.tex
\section{Related Work}
\label{sec:related work}

\subsection{Instruction-based Video Editing Methods}
Recently, IVE has attracted considerable attention, aiming to modify videos 
according to natural language commands. Due to the scarcity of paired training 
data, InsV2V~\cite{InsV2V} proposes synthesizing video triplets, a paradigm 
further scaled by InsViE~\cite{Insvie-1m}. AnyV2V~\cite{ku2024anyv2v} explores 
a training-free plug-and-play approach that decomposes editing into first-frame 
modification and I2V propagation. StableV2V~\cite{StableV2V} focuses on shape 
and temporal consistency, while Omni-Video~\cite{tan2025omnivideo} connects MLLM 
with video diffusion models for diverse video tasks. LucyEdit
leverages large-scale training to achieve high visual fidelity and temporal 
consistency. However, these methods largely focus on target single-task 
appearance editing and struggle with complex structural edits.

\subsection{Instruction-based Video Editing Datasets}
Video editing datasets often rely on existing video editing models for data 
generation. However, due to the inherent limitations of these models, 
constructing high-quality datasets urgently requires a rigorous and standardized 
filtering process. Existing datasets such as InsViE~\cite{Insvie-1m}, and Se\~norita-2M~\cite{zi2026senorita} either lack 
unified filtering criteria or employ insufficiently rigorous screening procedures, 
resulting in uneven data quality with residual static videos and failed editing samples.
Ditto~\cite{ditto} improves quality control but is heavily skewed toward style 
transfer, limiting task diversity. Concurrent OpenVE-3M~\cite{OpenVE} improves 
filtering rigor, yet its task scope remains confined to appearance-level edits, 
without support for structural or complex editing.

\begin{table*}[t]
\centering
\caption{A Detailed Comparison of \OurData with Prior and Concurrent IVE Datasets.}
\label{tab:data_compare}
\resizebox{\textwidth}{!}{%
\setlength{\tabcolsep}{4pt}
\fontsize{6}{8}\selectfont
\renewcommand{\arraystretch}{0.9}
\begin{tabular}{@{} l >{\columncolor{blue!5}}c ccccc @{}}
\toprule
\rowcolor{white}
Dimension & \OurData (Ours) & Ditto \cite{ditto} & Señorita-2M \cite{zi2026senorita} & InsViE \cite{Insvie-1m} & OpenVE-3M \cite{OpenVE} \\
\midrule
Dataset Scale    & 2M   & 1M      & 2M           & 1M    & 3M \\
Resolution       & 720p & 720p    &   336$\times$592\textasciitilde1120$\times$1984    & 576p  & 720p \\
Frames per Video & 65\textasciitilde129 & 101  & 33\textasciitilde64       & 25    & 65\textasciitilde129 \\
\midrule
Basic Edits      & \cmark & \cmark & \cmark      & \cmark & \cmark \\
Camera Movement  & \cmark & \xmark & \xmark      & \xmark & \cmark \\
Subject Movement & \cmark & \xmark & \xmark      & \xmark & \xmark \\
Reference-based Edit & \cmark & \xmark & \xmark  & \xmark & \xmark \\
Multi-Task Edit  & \cmark (2--5 tasks) & \xmark & \xmark & \xmark & \xmark \\
\midrule
MLLM      & Gemini2.5-Pro & Qwen-VL & Llama 3.2-8B & GPT-4o & GPT-4o \\
Benchmark & \cmark (\OurBench) & \xmark & \xmark & \xmark & \cmark (OpenVE-Bench) \\
\bottomrule
\end{tabular}
}
\end{table*}

%% file: 3_method.tex
\section{The \OurData Dataset}
\label{sec:dataset}

In this section, we present Goku, a large-scale and multifaceted dataset designed to push the boundaries of IVE beyond simplistic attribute modifications. 
While existing benchmarks primarily focus on single-task appearance-level edits, Goku is meticulously curated to encompass a diverse array of real-world challenges, including complex structural deformations, simultaneous multi-task interactions, and reference-guided editing.
The following subsections detail our systematic data collection pipeline (Sec.~\ref{sec:pipeline} and \ref{sec:appearance_synthesis}) and progressive filtering system (Sec.~\ref{sec:quality_assurance}) that together ensure high semantic fidelity and temporal consistency.
Due to space constraints, we provide more dataset statistics and visual results in our supplementary material.

\subsection{Video Pre-Processing}
\label{sec:pipeline}

\begin{figure}[ht]
  \centering
    \includegraphics[width=0.9\linewidth]{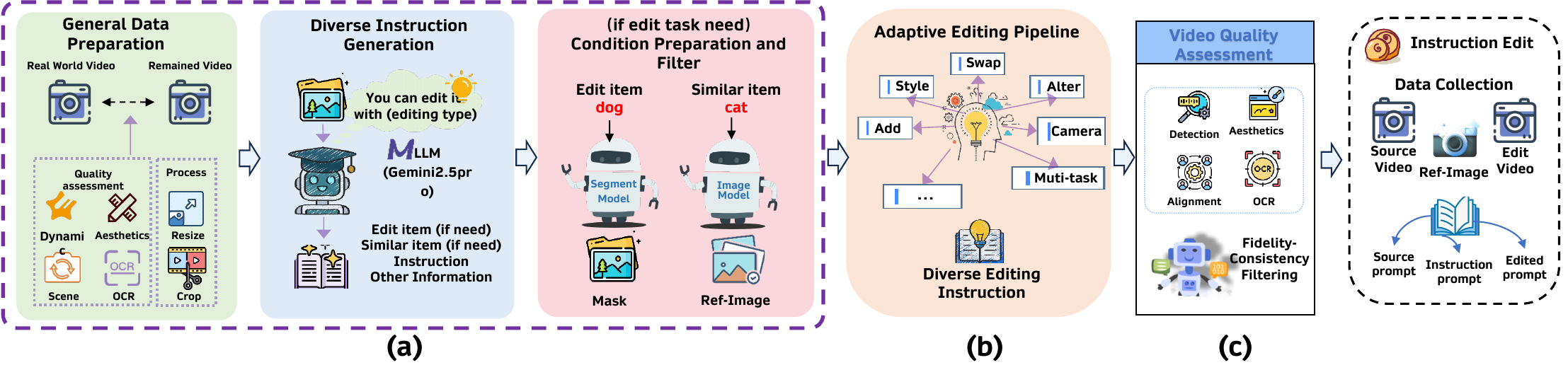}
    \caption{The illustration of our automated video editing pipeline. (a) Video Pre-Processing. (b) Data Generation for Different Tasks. (c) Progressive Filtering System.}
  \label{fig:DataPipeline}
\end{figure}

Starting from the raw video clips provided by Koala-36M~\cite{wang2025koala}, we develop a streamlined and automated pipeline to prepare the foundational components for our editing pairs (see Fig.~\ref{fig:DataPipeline}(a)).

\noindent\textbf{Quality Assessment and Filtering.}
We curate 1 million high-quality video clips from Koala-36M~\cite{wang2025koala}. The filtering pipeline applies shot transition detection, aesthetic scoring, motion dynamics analysis, and OCR-based watermark removal, followed by a content richness screening conducted by Gemini2.5-Pro. Each clip is trimmed to 3 to 10 seconds to highlight coherent actions and scenes.

\noindent\textbf{MLLM-Powered Instruction Generation.}
We leverage the multimodal understanding capabilities~\cite{liang2024hypercorrelation} of Gemini2.5-Pro to generate natural and diverse editing instructions for each task category. For Add, Remove, Swap, and Subject Movement tasks, the model first analyzes the video content and identifies operable target objects (covering over 200 categories), then outputs structured editing instructions along with object labels. Style Transfer tasks cover over 100 styles, and Camera Movement tasks cover over 20 motion patterns. For complex Multi-Task Editing and Structural Editing, we design dedicated prompts that guide Gemini2.5-Pro to analyze video content and identify the most suitable task categories, preventing task conflicts (e.g., simultaneously removing a dog and adding a hat on the dog's head). Detailed designs of the structured prompt templates are provided in our supplementary material.

\noindent\textbf{Reference Image/Mask Extraction.}
Based on the object labels produced in the previous stage, we employ Grounded-SAM2~\cite{ren2024groundedsam} to extract the temporal masks. To maximize the success rate and quality of the following VACE-based editing, we further use Gemini2.5-Pro to analyze the lighting conditions and object pose of the editing region. 
Then reference images of similar objects under the corresponding environment are generated, which serve as spatial and appearance conditions for subsequent synthesis.

\subsection{Data Generation for Different Tasks}
\noindent\textbf{Appearance Editing.}
\label{sec:appearance_synthesis}
Conventional instruction-based editing typically covers five appearance editing tasks: Add, Remove, Swap, Alter, and Style Transfer.

\noindent \textit{Add and Remove.}
Directly inserting objects into a video tends to produce artifacts such as floating objects and implausible occlusions. We circumvent this issue by exploiting the \textbf{duality} between Add and Remove.
Specifically, we first apply Minimax-Remover~\cite{zi2026minimax} to perform seamless removal of target objects, obtaining high-quality Remove samples; the roles of the original and removed videos are then swapped to yield natural Add samples.

\noindent \textit{Swap and Attribute Alter.}
We feed the source video, the temporal masks extracted by Grounded-SAM2, and the reference images generated by Flux jointly into VACE~\cite{jiang2025vace}, which performs semantic swap or attribute alter within the masked regions while preserving all non-edited areas.

\noindent \textit{Style Transfer.}
Applying style transfer directly to an entire video makes it difficult to maintain inter-frame consistency, often leading to flickering and style drift. Following the principle of \textbf{sub-problem decomposition}, we break this task into three controllable steps: (1) stylizing the first frame with Flux, (2) extracting per-frame depth maps as geometric constraints, and (3) feeding the stylized first frame with the depth sequence into VACE to propagate the style along the temporal axis. The depths ensure the scene's geometric structure remains unchanged during stylization, achieving temporally consistent style transfer.

\noindent \textbf{Structural Editing.}
\label{sec:structural_synthesis}
Unlike appearance editing that primarily alters textures or styles, structural editing involves the spatial shift or topological arrangement of the scene.
As existing datasets rarely address such dynamic transformations, we are the first to construct large-scale paired data for two representative categories: Subject Movement and Camera Movement.

\noindent \textit{Subject Movement.}
This task aims to modify the motion dynamics or spatial positioning of the primary subject within a video, rather than the appearance attributes. 
To tackle the inherent complexity of such transformations, we adopt a \textbf{sub-problem decomposition} principle, decoupling the task into two more manageable components: \textbf{action variation} and \textbf{position variation}.
For \textbf{action variation}, Gemini2.5-Pro is employed to analyze the source video and generate two distinct action descriptions for the same subject (e.g., ``walking'' $\rightarrow$ ``running'').
We then utilize Wan2.2 to synthesize videos conditioned on these descriptions, thereby obtaining high-quality paired data where the subject identity and background remain consistent while the actions diverge.
For \textbf{position variation}, we first address the relocation at the frame level by using Flux to shift the target object to a new spatial coordinate in the initial frame. Subsequently, we extend this relocated first frame into a temporally coherent video using Wan2.2.
This effectively reduces the challenge of video-level object relocation into two more tractable sub-problems: image-level editing and conditional video generation, significantly improving the data's structural diversity.

\noindent \textit{Camera Movement.}
Based on the instructions and scene descriptions generated by Gemini2.5-Pro, we use RecamMaster~\cite{bai2025recammaster} to synthesize camera movement video pairs. For complex camera motions (e.g., ``pan left then push in''), we similarly \textbf{decompose} them into sequential combinations of basic motion types which cover over 20 camera motion patterns.

\noindent \textbf{Multi-Task Editing.}
\label{sec:multi_synthesis}
Leveraging the task decomposition generated during the MLLM-Powered Instruction Generation stage, a single-task editing sequence is composed and executed step by step, with the output of each step serving as the input to the next. 
The overall quality and consistency of the editing are rigorously safe-guarded by our progressive filtering system (see Sec.~\ref{sec:quality_assurance}).

\noindent \textbf{Reference-based Editing.}
\label{sec:reference_synthesis}
Reference-based Editing (i.e., Reference Swap and Add) requires the model to perform edits guided by user-provided reference images.
Since the reference images are obtained by cropping the target region from the original video using masks, which are consistent with the target object, the model can degenerate into pixel-level copying.
To handle this, we employ Flux to repaint the reference images and provide perturbed reference images with pose offsets, lighting variations, and background replacements in our dataset.

\subsection{Progressive Filtering System}
\label{sec:quality_assurance}
To ensure the highest data quality and instruction-following precision, we implement a progressive filtering system consisting of three-tier critical quality gates embedded throughout our pipeline.

\noindent \textbf{Tier 1: Source Video Filtering.}
During the video pre-processing, we apply aesthetic scoring, motion dynamics analysis, shot transition detection, and OCR-based watermark removal to the raw videos from Koala-36M, followed by content richness screening conducted by Gemini2.5-Pro. This yields 1M high-quality source clips shared across all task categories.

\noindent \textbf{Tier 2: Condition Verification.}
Before data generation, we validate all intermediate representations to prevent error propagation.
Specifically, mask completeness is evaluated via Intersection over Union (IoU) thresholding.
Furthermore, we employ Gemini2.5-Pro to verify the semantic consistency between the detected editing targets and generated instructions, while also assessing the visual plausibility of synthesized reference images. Samples that fail any check are discarded, ensuring that only high-fidelity inputs enter the computationally expensive data synthesis stage.

\noindent \textbf{Tier 3: Post-Synthesis Validation.}
Each final editing pair undergoes a dual-level evaluation. At the low-level visual quality stage, we perform inter-frame consistency check, frequency-domain artifact detection, and aesthetic re-scoring. At the semantic level, Gemini2.5-Pro is utilized to assess editing accuracy and photorealism. This comprehensive validation is highly selective, filtering out approximately \textbf{88\%} of the synthesized samples to ensure the ultimate data quality.

To validate our progressive filtering system, we include a human evaluation (100 samples per task, 3 annotators) and precision/recall analysis in our supplementary material.
An alternative filtering pipeline based on open-source models (Qwen3VL-30B~\cite{bai2025qwen3}) is also provided for the reproducibility of our methodology.

\begin{figure}[t]
  \centering
    \includegraphics[width=0.9\linewidth]{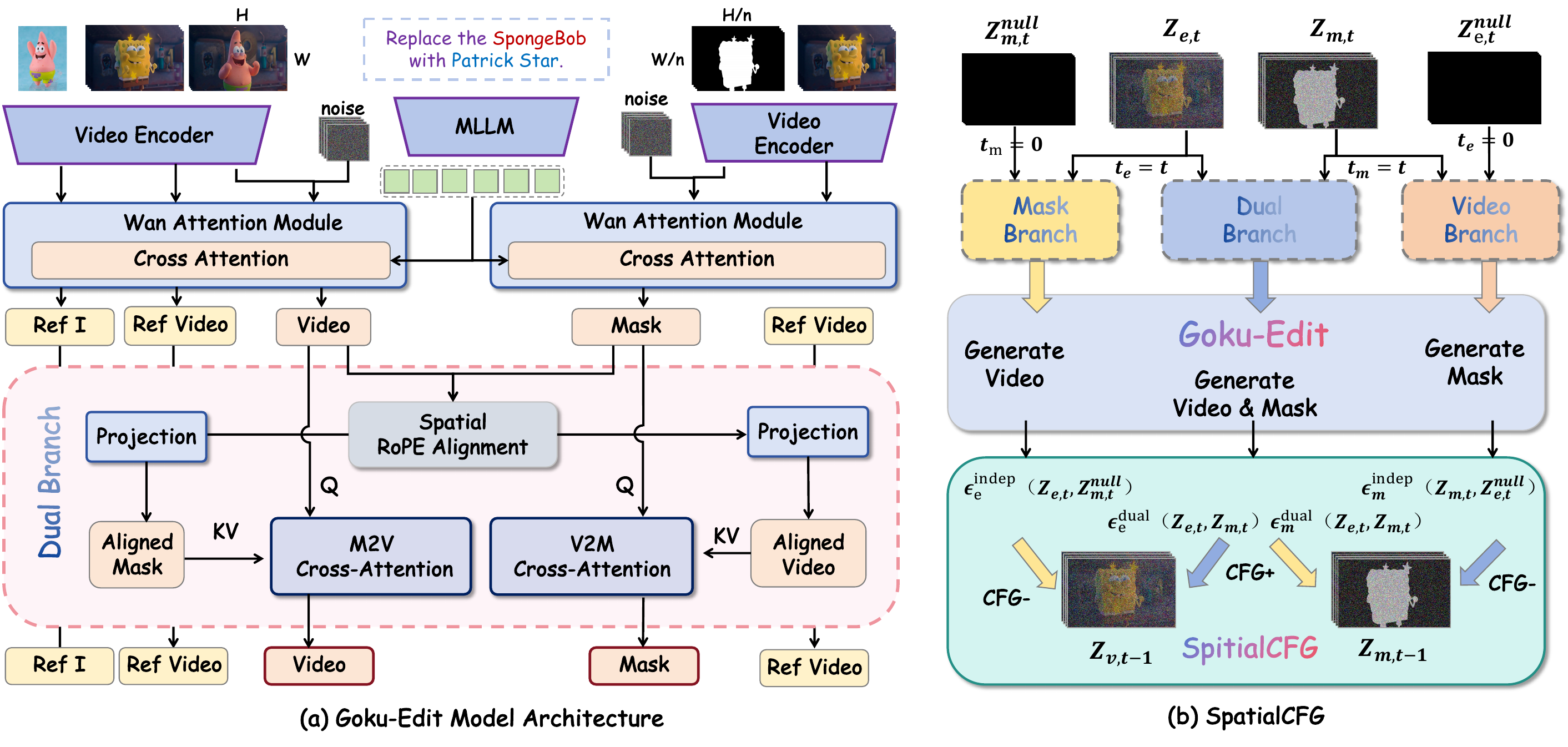}
\caption{Overview of \OurModel, featuring a dual-branch architecture with RoPE-aligned spatial cross-attention and inference-time SpatialCFG.}
  \label{fig:arc}
\end{figure}

\section{\OurModel Model}
\label{sec:model}
In this section, we present the design of \OurModel as illustrated in Figure~\ref{fig:arc}, comprising 
three core components. 
First, we introduce a \textbf{Dual-Branch Model 
Architecture} (Sec.~\ref{sec:dual_branch}) that couples a primary video 
 editing branch with an auxiliary mask-prediction branch, providing implicit 
spatial guidance for the generation process. 
Second, to resolve the spatial misalignment in cross-attention caused by the mask branch's lower operating resolution ($1/n$), we propose \textbf{RoPE-aligned spatial cross-attention} (Sec.~\ref{sec:rope_cross_attn}). This mechanism unifies positional embeddings into a shared physical coordinate system, ensuring precise cross-branch correspondence while maintaining computational efficiency.
Third, to further strengthen spatial constraints during inference, we design \textbf{Spatial Enhanced CFG} (Section~\ref{sec:spatial_cfg}).

\subsection{Dual-Branch Model Architecture}
\label{sec:dual_branch}

To achieve highly controllable video editing, we propose a dual-branch architecture where an auxiliary mask-prediction branch is introduced to provide spatial guidance for the main generation process. Both the primary video editing branch and the auxiliary mask branch adapt the pre-trained Wan2.2-5B model. Specifically, the primary video branch synthesizes the edited video $V_e$ conditioned on a text prompt $T_p$, the source video $V_s$, and a reference image $I_r$. Concurrently, the auxiliary mask branch predicts the mask $M$ of the targeted editing region, conditioned on $T_p$ and a spatially downsampled source video $V_d$ (subsampled by a factor of $n$ along spatial dimensions $H$ and $W$). To significantly enhance the comprehension of complex instructions, we utilize a  frozen Multimodal Large Language Model (Qwen3VL-8B~\cite{bai2025qwen3}) to process the text prompt $T_p$ for both branches. During the feature encoding phase, we employ a pre-trained VAE to encode $I_r$, $V_s$, $V_e$, $V_d$, and $M$ into their respective latent representations $z_r$, $z_s$, $z_e$, $z_d$, and $z_m$. In the denoising process, the conditioning latents are concatenated with the noisy target latents along the temporal dimension (denoted by $[\cdot, \cdot]_T$). Specifically, the composite input latents for the main video branch ($z'_{e,t}$) and the auxiliary mask branch ($z'_{m,t}$) are respectively formulated as:

\begin{equation}
\label{eq:composite_latents}
z'_{e,t} = [z_r, z_s, z_{e,t}]_T, \qquad z'_{m,t} = [z_d, z_{m,t}]_T
\end{equation}

These composite latents are then fed into their respective branches to predict the corresponding noise.

\subsection{RoPE-Aligned Spatial Cross-Attention}
\label{sec:rope_cross_attn}
The two branches interact via cross-attention, enabling the mask branch's structural control signals to guide the main branch's video generation. However, since the mask branch operates at $1/n$ spatial resolution, tokens from the two branches correspond to different discrete coordinate grids. In RoPE-based attention, the attention weight between a Query (position $j$) and a Key (position $k$) explicitly depends on the relative offset $j-k$. Performing cross-resolution attention directly introduces spurious macro-level offsets caused by grid mismatch, injecting unwanted rotational phase penalties into the feature inner product and inappropriately suppressing attention between spatially corresponding tokens.

\noindent\textbf{Spatial Alignment via RoPE Scaling.}
We address this by multiplying the mask branch's position indices by factor $n$ before RoPE computation, mapping a mask token at discrete index $(x, y)$ to high-resolution coordinate $(nx, ny)$ without modifying the RoPE frequency basis. After scaling, when a mask Key at $k'=(nx, ny)$ interacts with a video Query at $j=(nx+\delta_x, ny+\delta_y)$, where $\delta_x, \delta_y \in [0, n\text{-}1]$, the relative offset becomes: $j - k' = (\delta_x,\, \delta_y)$. Macro-level misalignment is thereby eliminated: spatially coincident positions ($\delta=0$) yield an identity rotation matrix ($R_0=I$), while the offsets of non-coincident token pairs remain proportional to their true physical distance, preserving RoPE's locality bias across resolutions.

\noindent\textbf{Bidirectional Cross-Attention.}
After aligning both branches to a unified RoPE coordinate system, we perform two cross-attention operations. In mask-to-video (M2V) attention, $z_e$ serves as Query and the aligned $z_m$ provides Key and Value, with the structural signal fused via a residual connection: 
\begin{equation}
    z_e^{\text{updated}} = z_e + \Delta z_e.
\end{equation}
Video-to-mask(V2M) attention reverses the roles; since the mask branch's indices have been scaled to the high-resolution system, both directions share the same alignment scheme, giving $z_m^{\text{updated}} = z_m + \Delta z_m$.
The reverse pass is crucial: without V2M, the mask branch predicts structure solely from the downsampled source video, unaware of the main branch's evolving state. As denoising progresses, the predicted mask diverges from the edited content, causing flickering and boundary incoherence. V2M closes this loop, enabling the mask branch to refine its predictions at each step based on the main branch's state.

\subsection{Spatial Enhanced CFG (SpatialCFG)}
\label{sec:spatial_cfg}

Standard text CFG only amplifies the text-conditioned signal and cannot further emphasize the spatial constraints from the mask branch, potentially causing boundary drifting and editing spillover in complex structural edits. We propose \textbf{SpatialCFG}, a training-free inference strategy that explicitly amplifies cross-branch spatial constraints by contrasting coupled and decoupled predictions.

When bidirectional cross-attention (M2V and V2M) is enabled, the model produces coupled predictions $\hat{\epsilon}_{e}^{\mathrm{dual}}(z'_{e,t}, z'_{m,t})$ and $\hat{\epsilon}_{m}^{\mathrm{dual}}(z'_{e,t}, z'_{m,t})$ for the video and mask branches, respectively. To isolate the spatial signal injected by each branch, we construct decoupled baselines by disabling cross-branch attention and replacing the counterpart input with a null latent: $\hat{\epsilon}_{e}^{\mathrm{indep}}(z'_{e,t}, z^{\mathrm{null}}_{m,t})$ for the video branch (M2V disabled) and $\hat{\epsilon}_{m}^{\mathrm{indep}}(z'_{m,t}, z^{\mathrm{null}}_{e,t})$ for the mask branch (V2M disabled). Both baselines retain text conditioning and intra-branch computation. SpatialCFG amplifies the cross-branch increment for each branch independently:
{\scriptsize
\begin{equation}
\begin{aligned}
\tilde{\epsilon}_e &= \hat{\epsilon}_{e}^{\mathrm{indep}}(z'_{e,t}, z^{\mathrm{null}}_{m,t})
+ s_e \Big(
  \hat{\epsilon}_{e}^{\mathrm{dual}}(z'_{e,t}, z'_{m,t})
  - \hat{\epsilon}_{e}^{\mathrm{indep}}(z'_{e,t}, z^{\mathrm{null}}_{m,t})
\Big), \\
\tilde{\epsilon}_m &= \hat{\epsilon}_{m}^{\mathrm{indep}}(z'_{m,t}, z^{\mathrm{null}}_{e,t})
+ s_m \Big(
  \hat{\epsilon}_{m}^{\mathrm{dual}}(z'_{e,t}, z'_{m,t})
  - \hat{\epsilon}_{m}^{\mathrm{indep}}(z'_{m,t}, z^{\mathrm{null}}_{e,t})
\Big),
\end{aligned}
\end{equation}}
where $s_e$ suppresses unintended modifications outside the editing region and $s_m$ refines mask boundary consistency. SpatialCFG composes orthogonally with standard text CFG: we perform the above under both text-conditioned and text-unconditioned settings to obtain $(\tilde{\epsilon}_e^{\mathrm{cond}}, \tilde{\epsilon}_m^{\mathrm{cond}})$ and $(\tilde{\epsilon}_e^{\mathrm{uncond}}, \tilde{\epsilon}_m^{\mathrm{uncond}})$, then apply text guidance to each branch separately:
{\scriptsize
\begin{equation}
\hat{\epsilon}_b = \tilde{\epsilon}_b^{\mathrm{uncond}} + s_{\mathrm{text}}\left(\tilde{\epsilon}_b^{\mathrm{cond}} - \tilde{\epsilon}_b^{\mathrm{uncond}}\right), \quad b \in \{e, m\}.
\end{equation}}

%% file: 3_5_bench.tex
\section{\OurBench}
\label{sec:bench}

\begin{figure*}[t]
  \centering
    \includegraphics[width=0.9\linewidth]{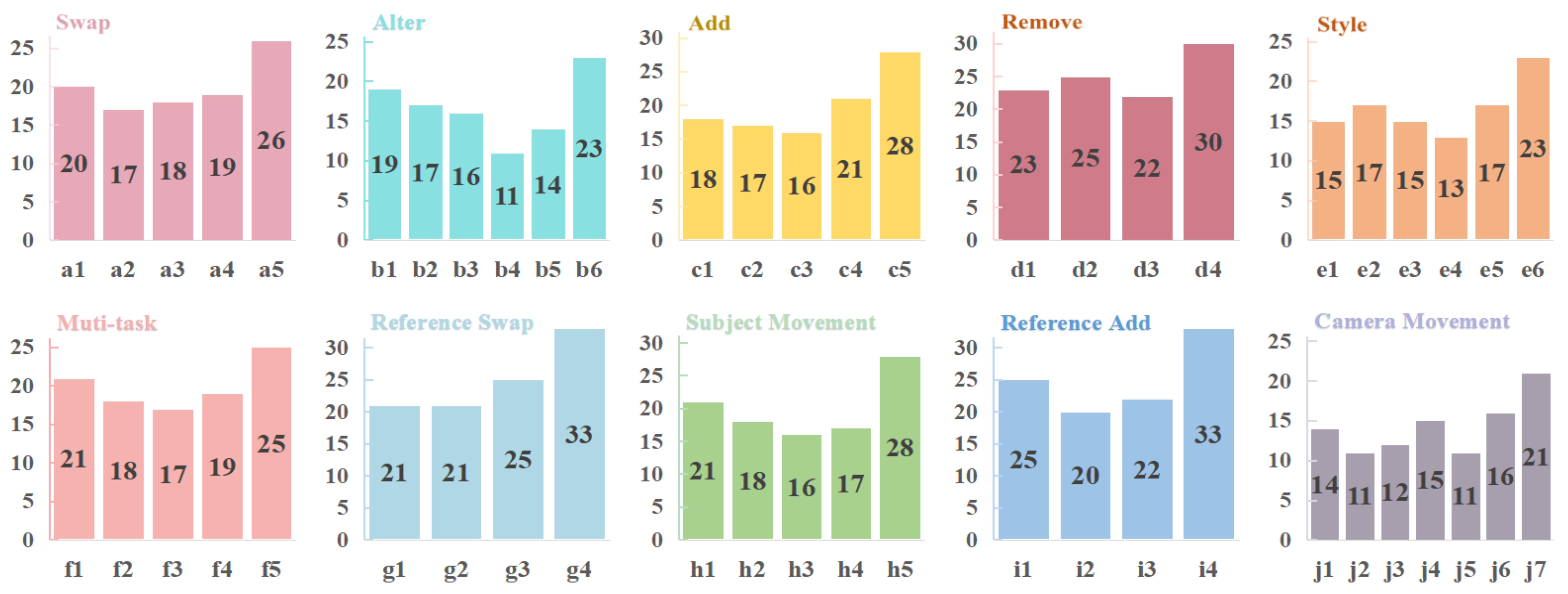}
    \caption{Statistical distributions of \OurBench.}
  \label{fig:leida}
\end{figure*}

Existing video editing benchmarks~\cite{ju2025editverse} are predominantly designed around single-task appearance editing, insufficient to comprehensively assess model capabilities on complex editing tasks (e.g., multi-task editing and structural editing).
To address this, in this section, we present \OurBench, a comprehensive video editing benchmark comprising 1,000 diverse test cases paired with \textbf{7 editing-specific metrics}, designed to provide fine-grained evaluation for complex editing capabilities along two dimensions: generation quality and instruction following. 
Specifically, we adopt general video quality metrics from VBench~\cite{huang2024vbench}, while our 7 custom metrics are meticulously designed to assess fine-grained editing fidelity and multi-task alignment, as detailed below.

\noindent\textbf{Test Set Construction.}
\OurBench curates 1,000 high-quality and challenging video clips from the Koala-36M dataset~\cite{wang2025koala}. Selection criteria include resolution ($\geq$720p), duration (3 to 10 seconds), and motion complexity, and so on; the complete selection pipeline and criteria are detailed in the supplementary material. The final test set covers multi-person scenarios, full and half body human subjects, animals (dogs, cats, sharks, birds, etc.), common objects (clothing, vehicles, buildings, etc.), and natural landscapes (mountains, rivers, deserts, etc.). Furthermore, we specifically include challenging filming conditions such as low lighting, fast moving subjects, and significant camera shake, and provide dedicated coverage of structured editing (e.g., subject movement) and multi-task joint editing, which are scenarios underrepresented in existing benchmarks, to ensure sufficient diversity across subject types, scene environments, and task difficulty. The editing instructions for each video are manually authored to ensure diversity, accuracy, and appropriate task difficulty. Detailed distribution statistics of editing task types are shown in Figure~\ref{fig:leida}.

\noindent \textbf{Editing-Specific Metric Design (7 Metrics).}
To measure key capabilities such as structured editing and complex instruction following, we propose 4 general editing metrics and 3 task-specific metrics. 
\textbf{Four general editing metrics} use Gemini2.5-Pro as the judging model, performing automated evaluation based on predefined scoring rubrics:
(1)~Physical Rule Fidelity (PR): evaluates whether the edited results conform to real-world physical laws in terms of motion and interaction, focusing on the plausibility of physical interactions; (2)~Spatial Relationship Accuracy (SR): evaluates whether the spatial arrangement between subjects and the surrounding scene strictly adheres to the editing instructions; (3)~Instruction Following (IF): comprehensively evaluates the completeness of the model's execution of complex multi-task editing instructions; (4)~Overall Editing Quality (EQ): provides a holistic assessment of the results from the perspectives of visual naturalness and editing consistency.
\textbf{For the task-specific metrics}:
(5)~SuM (Subject Motion): for subject motion editing, uses Gemini2.5-Pro to score the realism and fluency of the subject's motion trajectory. The distinction from PR is that PR evaluates the overall plausibility of all physical interactions in the edited result, whereas SuM focuses specifically on subject's motion trajectory;
(6)~CM (Camera Motion): for camera motion editing, employs optical flow analysis to identify and evaluate the motion type across video frames;
(7)~ST (Style Transfer): for style transfer tasks, computes the DINO~\cite{caron2021emerging} feature similarity between the reference style image corresponding to the style text and the generated video frames to quantify style transfer accuracy.


%% file: 4_experiment.tex
\section{Experiment}
\label{sec:experiment}

\subsection{Baselines}
Leveraging \OurBench, we conduct a comprehensive evaluation of representative 
instruction-based video editing methods. For open-source approaches, we 
benchmark against TokenFlow~\cite{qu2025tokenflow}, InsV2V~\cite{InsV2V}, 
StableV2V~\cite{StableV2V}, InsViE~\cite{Insvie-1m}, AnyV2V~\cite{ku2024anyv2v}, 
Omni-Video~\cite{tan2025omnivideo}, and LucyEdit. In terms of 
closed-source commercial models, we compare our method, 
Runway Gen-4, and Luma Ray3. To further validate the 
generalization capability of our approach, we perform additional evaluations on EditVerse-Bench~\cite{ju2025editverse}, with detailed results 
provided in the supplementary material.

\subsection{Comparison on \OurBench} We first conduct a comprehensive quantitative comparison between the \OurModel model and state-of-the-art video editing methods. The results shown in Table~\ref{tab:compare-all} that \OurModel achieves the best performance on most evaluation metrics. Meanwhile, \OurModel also excels in task-related metrics, further confirming the shortcomings of previous datasets that lacked task design. With regard to editing fidelity, the \OurModel model scores significantly higher than other methods, showcasing its superior capability in understanding and executing instructions. Notably, the model also exhibits substantial advantages in adhering to physical laws and spatial relationships, providing strong evidence for the high fidelity and rationality of the edited videos.
Although \OurModel falls behind commercial models on perceptual quality metrics such as CLIP, MS, and AES, it achieves notably higher PR, SR, CM, and SuM scores, confirming that our structured task definitions and dedicated data construction endow the model with stronger capabilities in physical plausibility, spatial reasoning, and compositional understanding. Meanwhile, the superior FVD, BC, and TC metrics demonstrate that our Progressive Filtering System effectively mitigates temporal artifacts and preserves background consistency.

\begin{table}[ht]
\centering
\caption{Quantitative comparison results with previous methods on \OurBench.}
\label{tab:compare-all}

\setlength{\tabcolsep}{3pt}
\renewcommand{\arraystretch}{1}

\resizebox{\textwidth}{!}{%
\begin{tabular}{l|c|ccccccc|ccc|cccc}
\toprule
\multicolumn{16}{c}{\textbf{Instruction Task}} \\
\toprule
\textbf{MODEL} & \makecell{\textbf{Open-}\\\textbf{Source}} & \textbf{SC}$\uparrow$ & \textbf{BC}$\uparrow$ & \textbf{CLIP}$\uparrow$ & \textbf{FVD} $\downarrow$& \textbf{TC} $\uparrow$& \textbf{MS} $\uparrow$& \textbf{AES} $\uparrow$& \textbf{ST}$\uparrow$ & \textbf{SuM}$\uparrow$ & \textbf{CM}$\uparrow$ & \textbf{PR}$\uparrow$ & \textbf{SR}$\uparrow$ & \textbf{IF}$\uparrow$ & \textbf{EQ}$\uparrow$ \\
\toprule

TokenFlow   & \checkmark & --- & 0.911 & 0.131 & 4539.82 & 0.899 & 0.94 & 0.42 & 0.514 & 0.426 & 0.457 & 0.34 & 0.63  & 0.44  & 0.32 \\
InsV2V      & \checkmark & --- & 0.915 & 0.122 & 3988.01 & 0.951 & 0.96 & 0.56 & 0.485 & 0.533 & 0.542 & 0.358 & 0.284 & 0.391 & 0.317 \\
StableV2V   & \checkmark & --- & 0.938 & 0.257 & 3129.58 & 0.921 & 0.97 & 0.45 & 0.642 & 0.545 & 0.631 & 0.297 & 0.331 & 0.375 & 0.304 \\
InsViE      & \checkmark & --- & 0.929 & 0.379 & 2314.89 & 0.953 & 1.08 & 0.47 & 0.535 & 0.438 & 0.597  & 0.382 & 0.273 & 0.349 & 0.361 \\
AnyV2V      & \checkmark & --- & 0.922 & 0.243 & 2876.93 & 0.915 & 0.93 & 0.39 & 0.598 & 0.598 & 0.494  & 0.312 & 0.366 & 0.259 & 0.388 \\
Omni-Video   & \checkmark & --- & 0.966 & 0.369 & 1032.08 & 0.947 & 1.03 & 0.43 & 0.614 & 0.597 & 0.481 & 0.58  & 0.631  & 0.51  & 0.59 \\
LucyEdit    & \checkmark & --- & 0.926 & 0.361 & 1420.36 & 0.954 & 0.95 & 0.51 & 0.694 & 0.598 & 0.637 & 0.476 & 0.755 & 0.549 & 0.579 \\
\rowcolor{blue!10} Ours        & \checkmark & --- & \textbf{0.969} & \textbf{0.432} & \textbf{993.93} & \textbf{0.955} & \textbf{1.15} & \textbf{0.59} & \textbf{0.955} & \textbf{0.633} & \textbf{0.927} & \textbf{0.738} & \textbf{0.832} & \textbf{0.627} & \textbf{0.645} \\

\midrule

\textcolor{gray}{\textit{Runway}} & \textcolor{gray}{\xmark} & \textcolor{gray}{---} & \textcolor{gray}{0.958} & \textcolor{gray}{0.472} & \textcolor{gray}{1038.52} & \textcolor{gray}{0.947} & \textcolor{gray}{1.33} & \textcolor{gray}{0.65} & \textcolor{gray}{0.968} & \textcolor{gray}{0.614} & \textcolor{gray}{0.891} & \textcolor{gray}{0.705} & \textcolor{gray}{0.793} & \textcolor{gray}{0.758} & \textcolor{gray}{0.782} \\
\textcolor{gray}{\textit{Luma}}   & \textcolor{gray}{\xmark} & \textcolor{gray}{---} & \textcolor{gray}{0.951} & \textcolor{gray}{0.461} & \textcolor{gray}{1095.64} & \textcolor{gray}{0.940} & \textcolor{gray}{1.29} & \textcolor{gray}{0.63} & \textcolor{gray}{0.957} & \textcolor{gray}{0.601} & \textcolor{gray}{0.872} & \textcolor{gray}{0.681} & \textcolor{gray}{0.769} & \textcolor{gray}{0.741} & \textcolor{gray}{0.761} \\

\toprule
\multicolumn{16}{c}{\textbf{Reference Image + Instruction Task}} \\
\toprule

StableV2V & \checkmark & 0.41 & 0.931 & 0.382 & 2401.55 & 0.948 & 1.06 & 0.45 & --- & --- & --- & 0.491 & 0.762 & 0.558 & 0.583 \\
AnyV2V    & \checkmark & 0.45 & 0.925 & 0.255 & 2750.11 & 0.919    & 0.94  & 0.41  & --- & --- & --- & 0.533 & 0.78 & 0.572 & 0.591 \\
\rowcolor{blue!10} Ours      & \checkmark & \textbf{0.54} & \textbf{0.968} & \textbf{0.417} & \textbf{925.55} & \textbf{0.958} & \textbf{1.17} & \textbf{0.52} & --- & --- & --- & \textbf{0.718} & \textbf{0.832} & \textbf{0.851} & \textbf{0.824} \\

\midrule

\textcolor{gray}{\textit{Runway}} & \textcolor{gray}{\xmark} & \textcolor{gray}{0.58} & \textcolor{gray}{0.954} & \textcolor{gray}{0.465} & \textcolor{gray}{1028.73} & \textcolor{gray}{0.948} & \textcolor{gray}{1.32} & \textcolor{gray}{0.63} & \textcolor{gray}{---} & \textcolor{gray}{---} & \textcolor{gray}{---} & \textcolor{gray}{0.691} & \textcolor{gray}{0.802} & \textcolor{gray}{0.872} & \textcolor{gray}{0.848} \\
\textcolor{gray}{\textit{Luma}}   & \textcolor{gray}{\xmark} & \textcolor{gray}{0.55} & \textcolor{gray}{0.947} & \textcolor{gray}{0.453} & \textcolor{gray}{1081.29} & \textcolor{gray}{0.943} & \textcolor{gray}{1.28} & \textcolor{gray}{0.61} & \textcolor{gray}{---} & \textcolor{gray}{---} & \textcolor{gray}{---} & \textcolor{gray}{0.668} & \textcolor{gray}{0.785} & \textcolor{gray}{0.858} & \textcolor{gray}{0.831} \\

\bottomrule
\end{tabular}%
}

\end{table}

\begin{figure}[t]
\centering
\begin{adjustbox}{valign=t,minipage=0.38\linewidth}
  \centering
  \includegraphics[width=0.85\linewidth]{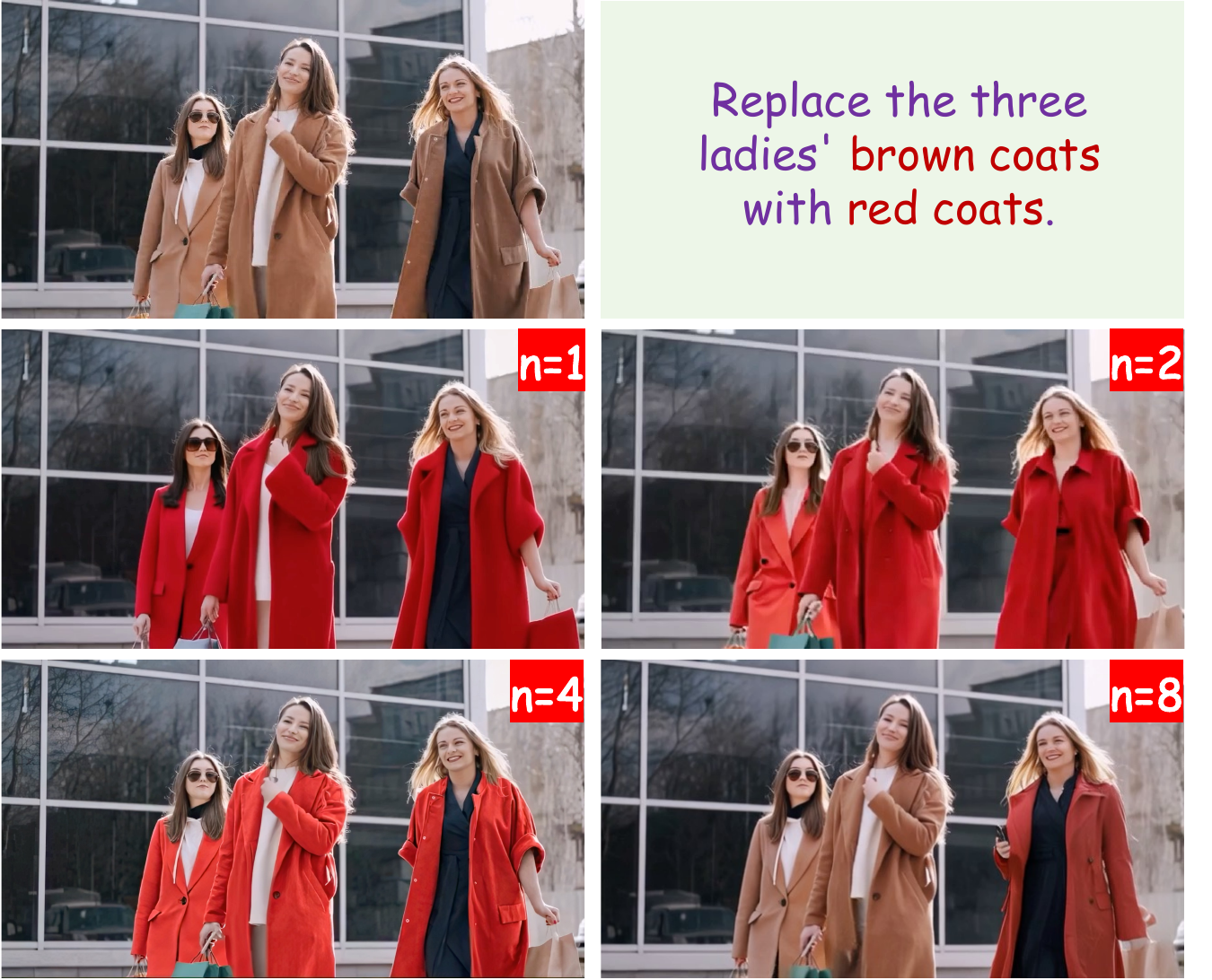}
  \captionof{figure}{Ablation study on the spatial downsampling factor $n$.}
  \label{fig:ablation-n}
\end{adjustbox}
\hfill
\begin{adjustbox}{valign=t,minipage=0.56\linewidth}
  \centering
  \setlength{\tabcolsep}{3pt}
  \renewcommand{\arraystretch}{1}
  \fontsize{6}{7}\selectfont
  \captionof{table}{Ablation study on training data. Models are evaluated 
  on \OurBench. \colorbox{blue!10}{Blue} row denotes the best configuration.}
  \begin{tabularx}{\linewidth}{@{}lcccccr@{}}
    \toprule
    \textbf{Datasets} & \textbf{Nums} & \textbf{CLIP}$\uparrow$ & \textbf{FVD}$\downarrow$ & \textbf{TC}$\uparrow$ & \textbf{IF}$\uparrow$ & \textbf{EQ}$\uparrow$ \\
    \midrule
    InsV2V     & 50k  & 0.3421 & 3102.15 & 0.891 & 0.298 & 0.241 \\
    InsViE     & 50k  & 0.3398 & 2843.67 & 0.873 & 0.301 & 0.253 \\
    Se\~norita & 50k  & 0.3502 & 2761.44 & 0.868 & 0.289 & 0.278 \\
    \midrule
    Ours (w/o) & 50k  & 0.3541 & 2512.08 & 0.862 & 0.318 & 0.331 \\
    Ours       & 50k  & 0.3780 & 1380.45 & 0.881 & 0.501 & 0.522 \\
    \rowcolor{blue!10}
    Ours       & 100k & 0.3891 & 1241.33 & 0.893 & 0.519 & 0.538 \\
    \bottomrule
  \end{tabularx}
  \label{tab:ablation-multitask}
\end{adjustbox}
\end{figure}

\subsection{Qualitative Evaluation}

\begin{figure}[h]
  \centering
    \includegraphics[width=0.95\linewidth]{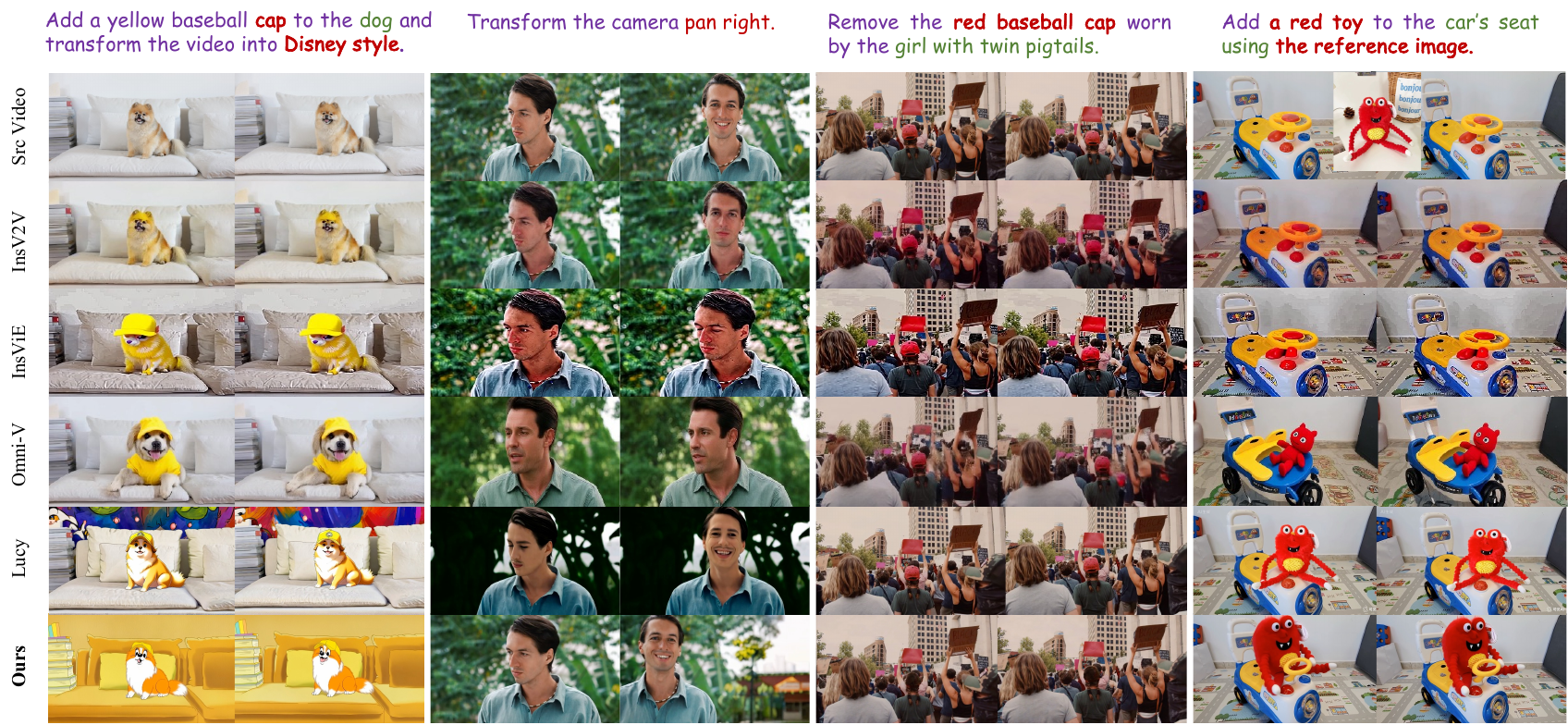}
    \caption{Comparison with state-of-the-art methods. }
  \label{fig:Compare3}
\end{figure}

Due to the limitations of quantitative metrics in evaluating editing tasks, we further validate the effectiveness of our approach through qualitative comparisons, as shown in Figure~\ref{fig:Compare3}.
\textbf{1) For structural editing tasks}, as shown in the second column, given the
instruction to pan the camera rightward, InsV2V, InsViE, Omni-Video, and
LucyEdit all fail to produce any effective camera motion, whereas \OurModel
correctly executes the pan operation and plausibly inpaints the newly
revealed regions. As shown in the third column, given the instruction to
remove the red hat from the girl's head in a complex scene, all baseline
methods fail to accurately localize the target region and introduce
varying degrees of damage to the surrounding content, while \OurModel
precisely localizes and completely removes the target while preserving
the background intact.
\textbf{2) For complex editing tasks}, as shown in the first column, given
the compound instruction to simultaneously add a yellow baseball cap to
the dog and transform the video into Disney style, all baseline methods
struggle to complete both edits simultaneously. LucyEdit manages to add the
cap but produces poor stylization results and damages the background,
whereas \OurModel accurately fulfills all editing instructions while
maintaining overall video quality.
\textbf{3) For reference-based editing tasks}, as shown in the fourth column,
baseline methods generally suffer from inaccurate target localization
and poor consistency with the reference image, while \OurModel precisely
identifies the placement location and maintains high visual consistency
with the reference appearance, significantly outperforming all
competing methods.

\newcolumntype{Y}{>{\centering\arraybackslash}X}

\begin{table}[t]
\noindent
\begin{minipage}[t]{0.58\linewidth}
\captionof{table}{Ablation study on the core components.}
\label{tab:ablationcore}
\setlength{\tabcolsep}{2pt}
\renewcommand{\arraystretch}{1}
\fontsize{6}{7}\selectfont

\begin{tabularx}{\linewidth}{cccc|YYYY}
\toprule
MLLM & Dual & RoPE & SpatialCFG & PR $\uparrow$ & SR $\uparrow$ & IF $\uparrow$ & EQ $\uparrow$ \\
\midrule
     &      &      &            & 0.651 & 0.743 & 0.541 & 0.578 \\
\checkmark &      &      &            & 0.664 & 0.756 & 0.573 & 0.591 \\
\checkmark & \checkmark &      &        & 0.695 & 0.789 & 0.588 & 0.612 \\
\checkmark & \checkmark & \checkmark &  & 0.718 & 0.819 & 0.608 & 0.631 \\
\rowcolor{blue!10}
\checkmark & \checkmark & \checkmark & \checkmark & \textbf{0.738} & \textbf{0.832} & \textbf{0.627} & \textbf{0.645} \\
\bottomrule
\end{tabularx}
\end{minipage}%
\hspace{0.6em} 
\begin{minipage}[t]{0.40\linewidth}

\setlength{\tabcolsep}{2pt}
\renewcommand{\arraystretch}{0.9}
\fontsize{6}{7}\selectfont

\makebox[\linewidth][r]{%
  \begin{minipage}[t]{\linewidth}
    \captionsetup{justification=raggedright,singlelinecheck=false} 
    \captionof{table}{User Study Results.}
    \label{tab:human_eval_results}

    \begin{tabularx}{\linewidth}{lYYY}
      \toprule
      \textbf{Method} & \textbf{IF $\uparrow$} & \textbf{VQ $\uparrow$} & \textbf{TC $\uparrow$} \\
      \midrule
      InsV2V \cite{InsV2V}              & 3.30 & 3.10 & 3.18 \\
      InsViE \cite{Insvie-1m}           & 3.42 & 3.18 & 3.25 \\
      Omni-Video \cite{tan2025omnivideo} & 3.75 & 3.82 & 3.90 \\
      LucyEdit          & 4.05 & 4.12 & 4.08 \\
      \midrule
      \rowcolor{blue!10} \OurModel      & \textbf{4.58} & \textbf{4.51} & \textbf{4.65} \\
      \bottomrule
    \end{tabularx}
  \end{minipage}%
}
\end{minipage}

\end{table}

\subsection{Ablation Studies}
\noindent\textbf{Data ablation on multi-task editing with \OurData.}
To verify the data contribution of \OurData on complex editing tasks, we compare models trained on different datasets under multi-task editing. For fair comparison, all methods adopt the same network architecture (LucyEdit), training scale (50k samples, 6k steps), and evaluation benchmark (\OurBench). As shown in Table~\ref{tab:ablation-multitask}, existing datasets primarily focus on single-task appearance editing, exhibiting limited performance on multi-task scenarios with consistently low IF and EQ scores. The model trained on unfiltered \OurData (Ours w/o) already surpasses all competing datasets, demonstrating that the expanded task coverage of \OurData alone yields substantial gains. After applying our Progressive Filtering System (Ours), IF and EQ improve further by a large margin, validating the effectiveness of data quality control for complex editing tasks. Scaling to 100k training samples yields consistent improvements across all metrics, indicating that performance gains from \OurData remain unsaturated and further scaling continues to be beneficial.

\noindent\textbf{Ablation study on the core components of \OurModel.}
Table~\ref{tab:ablationcore} progressively validates each core component. Replacing the text encoder with an MLLM improves IF from 0.541 to 0.573, indicating stronger semantic understanding of complex editing instructions. Incorporating the dual-branch architecture further improves PR and SR by 0.031 and 0.033, confirming that decoupling structural control from appearance rendering frees up the main branch's modeling capacity. Without RoPE alignment, cross-resolution spatial misalignment undermines the mask branch's guidance; after introducing RoPE-aligned spatial cross-attention, SR improves from 0.789 to 0.819, demonstrating that accurate cross-branch positional correspondence is essential for boundary precision. Finally, SpatialCFG yields the best performance across all metrics, with notable gains in PR (+0.020) and IF (+0.019), showing that it effectively suppresses editing spillover and enhances instruction following without additional training overhead. 

\noindent\textbf{Ablation study on the spatial downsampling factor $n$.}
As shown in Figure~\ref{fig:ablation-n}, $n=1$ yields smooth but textureless 
garments due to insufficient decoupling between the two branches.
Increasing $n$ progressively restores fine-grained appearance detail, 
with $n=4$ producing complete texture and precise localization.
At $n=8$, the structural signal degrades, causing localization failure.
$n=4$ is adopted as our default configuration. Full quantitative results across all metrics for different values of n are reported in the supplementary material.

\noindent\textbf{User Study}
We conduct a user study where 30 participants each rate 100 videos on a 5-point Likert scale across three dimensions. As shown in Table~\ref{tab:human_eval_results}, \OurModel consistently ranks first, with particularly notable advantages in preserving non-edited regions and handling complex motion instructions.

%% file: 5_summary.tex
\section{Conclusion}
\label{sec:conclusion}

In this work, we address the key challenges hindering progress in instruction-based video editing by introducing \OurData, a large-scale, high-fidelity dataset that comprehensively covers diverse editing patterns, including camera movement, subject movement, reference edit, and multi-task patterns. Our rigorous data creation pipeline ensures exceptional quality and diversity, effectively overcoming common issues found in existing datasets, such as static content, semantic mismatches, and poor visual quality. Complementing this resource, we present \OurBench, a standardized benchmark with novel evaluation metrics, facilitating robust assessment of model performance across technical quality, semantic alignment, and temporal coherence. Empirical studies demonstrate that models trained on \OurData consistently achieve superior results on multiple tasks and metrics, significantly outperforming previous datasets and state-of-the-art methods.

\section*{Acknowledgments}
This work was supported in part by NSFC under Grant 62371434 and U25B2010, the Postdoctoral Fellowship Program of CPSF under Grant Number GZC20252293, the China Postdoctoral Science Foundation-Anhui Joint Support Program under Grant Number 2024T017AH, China Postdoctoral Science Foundation under Grant Number  2025M783529, Anhui Postdoctoral Scientific Research Program Foundation (No.2025A1015), the Fundamental Research Funds for the Central Universities(No. WK2100250064), ZGCA Project-C20250302.